\pdfoutput=1

\documentclass[11pt]{article}

\usepackage[final]{acl}

\usepackage{times}
\usepackage{latexsym}
\usepackage{graphicx}
\usepackage{enumitem}
\usepackage{todonotes}
\usepackage{subfigure}
\usepackage{subcaption}
\usepackage{amsmath}

\usepackage[T1]{fontenc}

\usepackage[utf8]{inputenc}

\usepackage{microtype}

\usepackage{inconsolata}

\title{On the Challenges of Creating Datasets for Analyzing Commercial Sex Advertisements to Assess Human Trafficking Risk and Organized Activity}

\author{Pablo Rivas$^{1*}$ \quad Tomas Cerny$^2$ \quad Alejandro Rodriguez Perez$^1$ \\ \bf Javier S Turek$^3$ \quad Laurie Giddens$^4$ \quad Gisela Bichler$^5$ \quad Stacie Petter$^6$ \\
  $^1$Baylor University \quad $^2$The University of Arizona \quad $^3$Intel Labs  \quad $^4$ University of North Texas \\ $^5$California State University San Bernardino  \quad $^6$ Wake Forest University\\
  $^*$\texttt{Pablo\_Rivas@Baylor.edu}}
  
\begin{document}
\maketitle

\begin{abstract}
Our study addresses the challenges of building datasets to understand the risks associated with organized activities and human trafficking through commercial sex advertisements. These challenges include data scarcity, rapid obsolescence, and privacy concerns. Traditional approaches, which are not automated and are difficult to reproduce, fall short in addressing these issues. We have developed a reproducible and automated methodology to analyze five million advertisements. In the process, we identified further challenges in dataset creation within this sensitive domain. This paper presents a streamlined methodology to assist researchers in constructing effective datasets for combating organized crime, allowing them to focus on advancing detection technologies.
\end{abstract}

\section{Introduction}

The landscape of commercial sex advertisements is not just a platform for services but also a lucrative target for human traffickers and organized crime to exploit for financial gain. For law enforcement, the challenge is monumental – the volume and ever-renewing stream of ads make it almost impossible to keep up~\cite{giddens2023navigating}. With the evolution of NLP, there is a promising path forward to aid in identifying these suspicious ads; however, current approaches hinge on the availability and reliability of the datasets, lacking automation and reproducibility~\cite{10.1145/3563040}.

\begin{figure}[t!]
    \centering
    \includegraphics[width=20.5em]{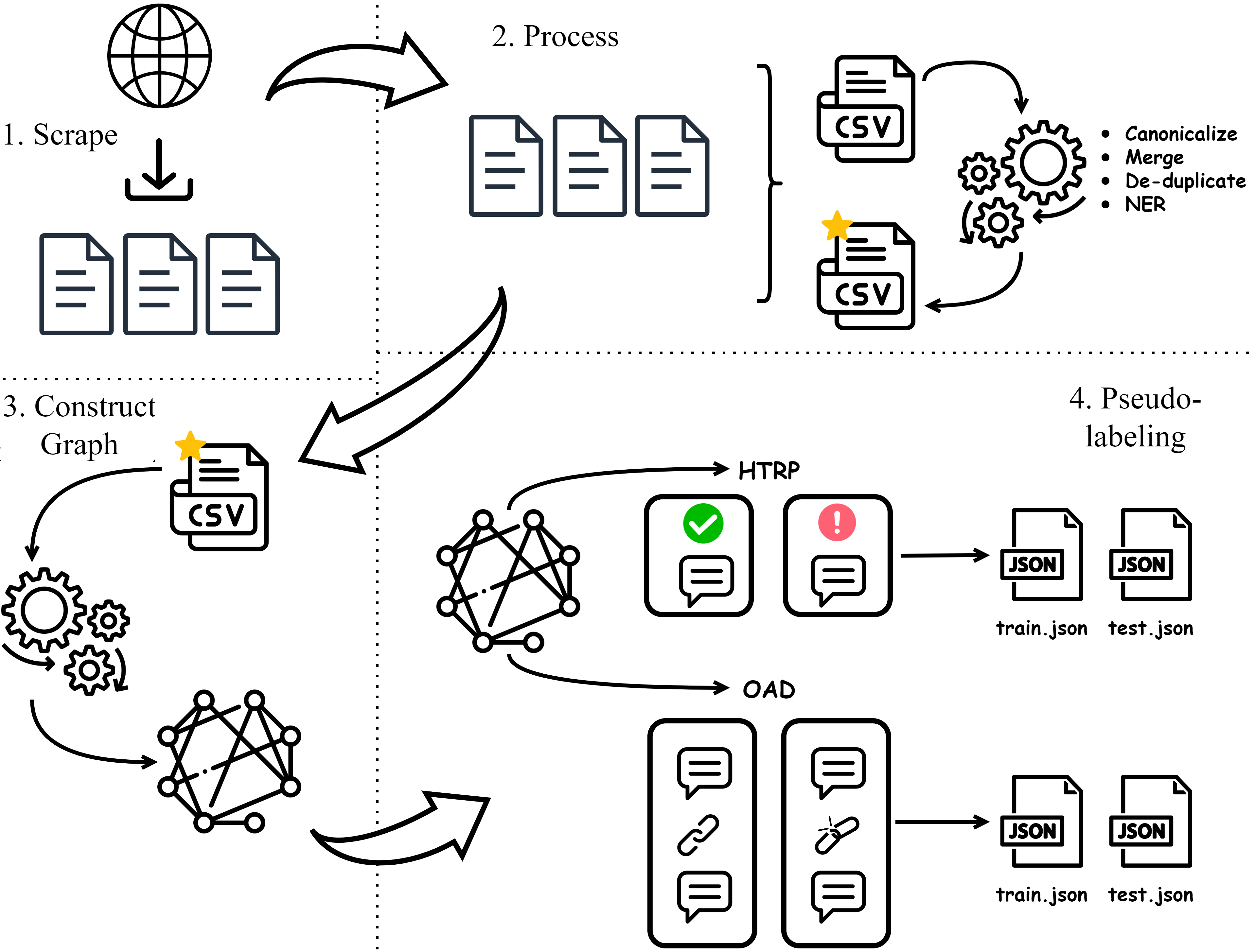}
    \caption{Methodology to generate a pseudo-labeled dataset in human trafficking risk prediction and organized activity detection tasks.}
    \label{fig:datamethod}
\end{figure}

In our pursuit to bolster the efforts of criminal investigators in detecting illicit activities, we embarked on a two-year journey to compile a comprehensive dataset of commercial sex ads, using the methodology depicted in Figure~\ref{fig:datamethod}. This endeavor initially appeared straightforward and unveiled various unexpected hurdles and obstacles. We intend not to showcase the dataset but to share the lessons learned when creating it. Due to the rapid pace at which this data can become obsolete, we focus on the methodology behind building this dataset. We aim to illuminate the challenges and pitfalls encountered along the way, guiding fellow researchers. This insight will enable others to sidestep these challenges and more efficiently contribute to the collective fight against online crime.

\section{Paucity of Datasets in This Domain}

Securing data for human trafficking (HT) research is notoriously challenging, requiring the coordinated efforts of academics, law enforcement, and, at times, victims themselves. This collaboration is complex and hard to manage. Innovative approaches, including using heuristics as stand-ins for direct indicators of trafficking and semi-automated labeling, have helped researchers sidestep these issues. Yet, accessing even the most basic raw data has proven difficult, demanding significant effort to extract from public sources, often with limited outcomes \cite{dubrawski2015leveraging, portnoff2017backpage, hundman2018always}. 

Language is a crucial tool in online illicit activities, including HT. Criminals disguise their activities and intentions through coded messages, crafting a specialized language that evolves with cultural and societal shifts. This evolution complicates research, as traffickers continually alter their communication to evade detection, making any model based on static data quickly outdated \cite{dubrawski2015leveraging, latonero2011human}. Presenting a solid, reproducible approach for collecting, processing, and labeling data from commercial sex ads will significantly bolster the creation of current datasets.

\section{Methodology Overview}

Embarking on a project to identify Human Trafficking Risk Prediction (HTRP) and Organized Activity Detection (OAD) from commercial sex advertisements, we devised a methodology demanding minimal human oversight. Illustrated in Figure~\ref{fig:datamethod}, our approach navigates through data scraping, processing, and analysis, leading to a graph-based model to identify risk of trafficking and organized crime.

Initially, we scrape a website for commercial sex to collect ads and metadata, uniformizing data. This process involves data cleaning and normalization, addressing the challenge of ensuring that vital contextual cues like slang or emojis are not lost, contrary to conventional text processing practices \cite{zhu2019detecting,wiriyakun2022extracting}. To manage data diversity and address the issue of data duplication—where our analysis found a staggering 90\% rate of textual duplication—we applied deduplication techniques, refining our dataset to 515,865 unique ads out of an original total of 5,053,249 ads~\cite{rodriguez2023combatting}.

\begin{figure}[t]
    \centering
    \includegraphics[width=20.5em]{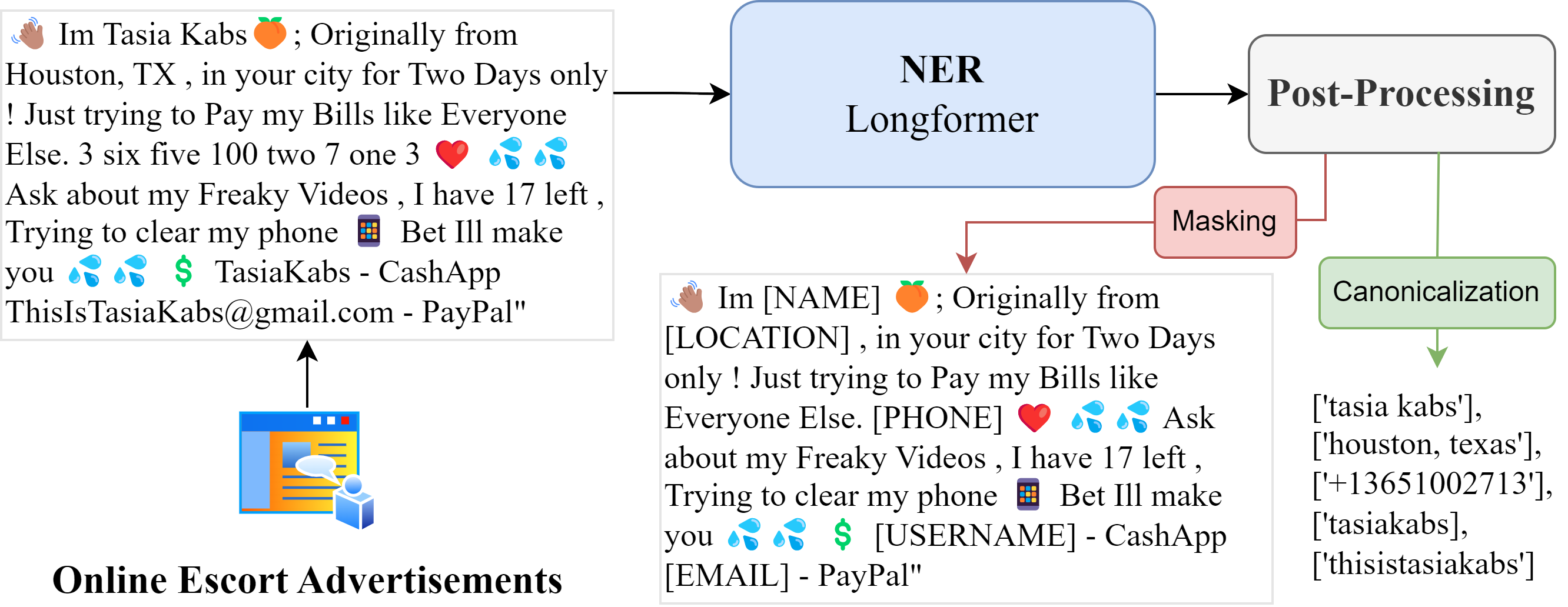}
    \caption{Processing the text of an ad with the NER pipeline. \emph{Personally identifiable data has been changed.}}
    \label{fig:ner_overall}
\end{figure}

For Named Entity Recognition (NER), illustrated in Figure~\ref{fig:ner_overall}, we encountered limitations with standard models. These models faltered against the adversarial text common in such ads, prompting us to custom-train NER models on a dataset of 1,810 labeled ads, identifying entities relevant to our investigation. Our exploration of various NER models highlighted the superiority of Longformer and XLNet, with Longformer slightly edging out due to its handling of out-of-vocabulary tokens.

\begin{figure*}
    \centering
    \begin{tabular}{cc}
        \includegraphics[width=0.49\linewidth]{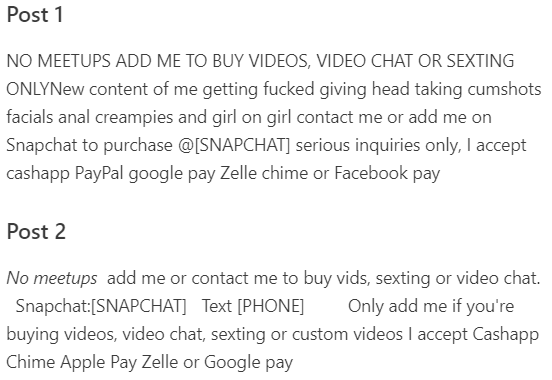} &
        \includegraphics[width=0.49\linewidth]{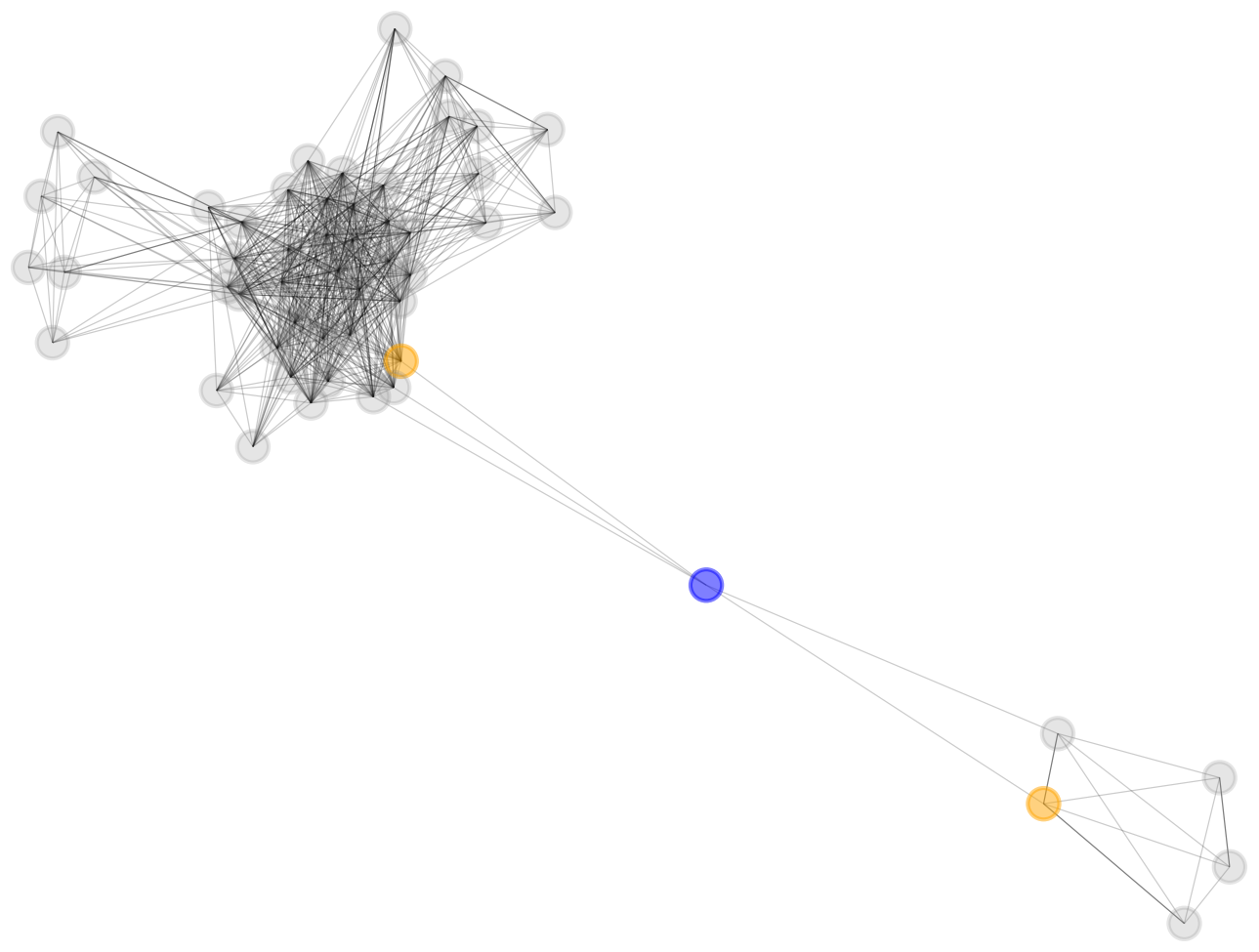} \\
        (a) & (b) \\
    \end{tabular}
    \caption{Ads connected indirectly in a connected component. (a) Description text of the highlighted posts. \emph{Personally identifiable data has been changed.} (b) Connected component graph where referred posts are in orange. }
    \label{fig:cc_power}
\end{figure*}

Constructing a Relatedness Graph from the deduplicated posts allowed us to visualize connections between ads through shared identifiers, like phone numbers or social media handles. Despite the vast potential connections, the graph revealed a sparse structure dominated by isolated nodes and significant connected components indicative of organized activities, as shown in Figure~\ref{fig:cc_power} and in Table~\ref{tab:cc_size}.

\begin{table}[t]
    \centering
    \caption{Size of the connected components.}
    \begin{tabular}{|r|r|}
        \hline
         \textbf{Size range} &  \textbf{Components} \\
         \hline
            1 node & 184,877\\
            2-10 nodes & 51,117\\
            10-100 nodes & 5,928\\
            100-1000 nodes & 80\\
            1000+ nodes & 1\\
            \hline
            \textbf{Total} & \textbf{287,192} \\ 
        \hline
    \end{tabular}
    
    \label{tab:cc_size}
\end{table}

Pseudo-labeling for OAD and HTRP involved leveraging the Relatedness Graph to classify ad pairs and individual ads, respectively, into binary categories. The division of this graph into connected components and their subsequent distribution into training and test sets underscored the complexities inherent in dataset preparation, especially given the unbalanced nature of real-world data. Our methodical approach to OAD involved the binary labeling of ad pairs based on their interconnectedness, emphasizing the importance of a balanced dataset and the necessity to discard excessively similar advertisements to mitigate bias. This was operationalized through a similarity threshold, informed by the Levenshtein distance, set at 0.5 after rigorous evaluation. For HTRP, we ventured beyond mere text analysis, employing heuristics based on ad metadata—such as the distance between locations exceeding 300 miles and the number of unique identifiers—to infer trafficking risk. 

\section{Limitations}

Our research encountered a few important limitations. Sparse data, NER pipeline errors, and data source heterogeneity introduce significant variability and potential inaccuracies. The process's susceptibility to bias—evidenced by differences in labeling driven by the algorithm's reliance on hard identifiers and geographical heuristics—raises concerns, as confirmed by statistical testing \cite{wilcoxon1945individual}; more specifically, a Wilcoxon rank-signed test with a $p$-value of $0.004$. These limitations, underscored by the variance in data treatment and the subjective selection of similarity metrics, highlight the complexities of deploying NLP techniques in the sensitive context of human trafficking, suggesting that while our approach marks a step forward, it navigates a landscape riddled with challenges.

\section{Discussion of Challenges}

\begin{figure}[t]
    \centering
    \includegraphics[width=0.83\columnwidth]{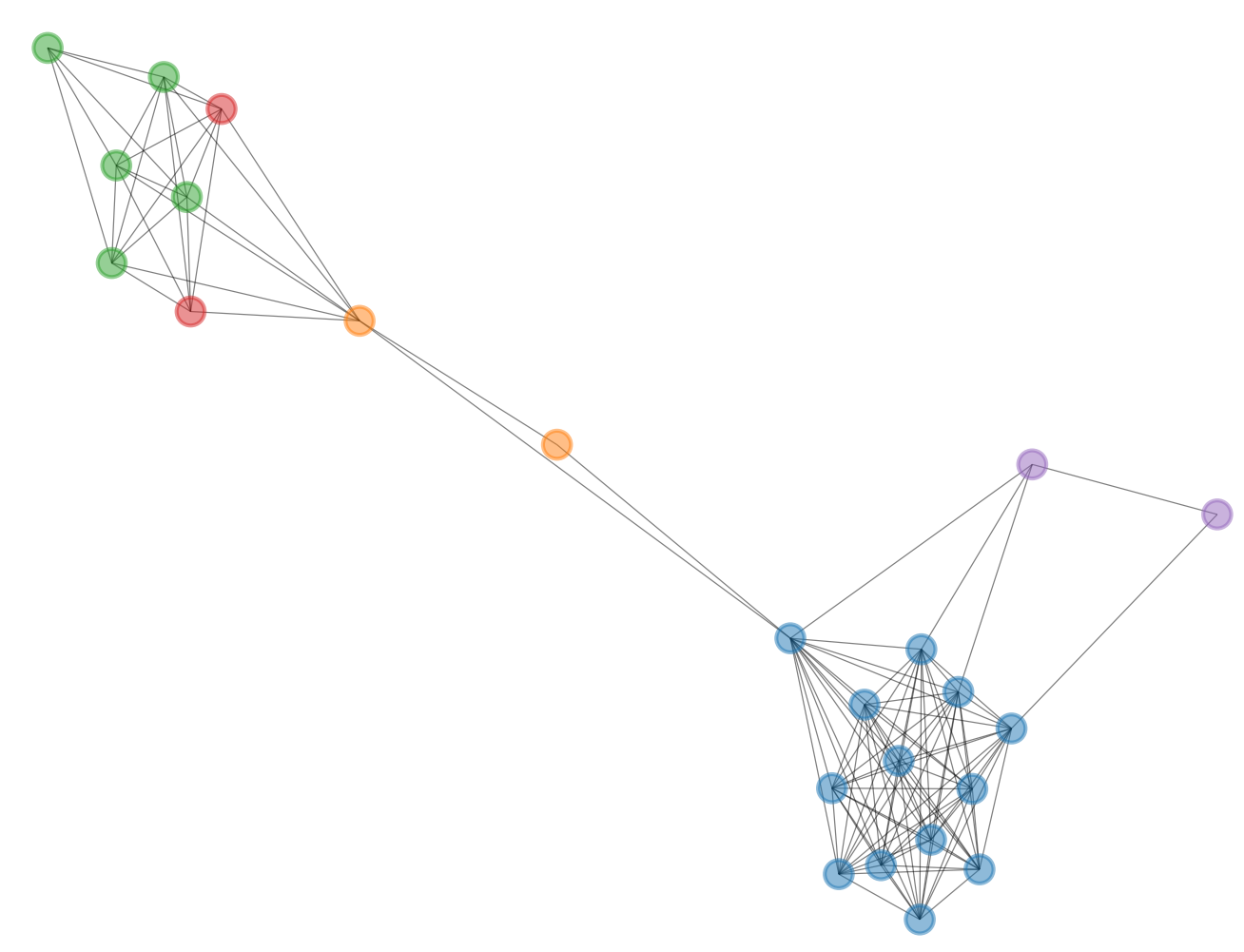}
    \caption{A connected component labeled positive due to several phone numbers found. Each color represents a different phone number encountered.}
    \label{fig:cc_identifiers}
\end{figure}

In constructing our dataset, we encountered critical challenges that underscore the complexity of this endeavor. First, the data's heterogeneous nature often meant that while some fields were missing, crucial information could still be embedded within text or images, necessitating a nuanced approach to extraction and interpretation. The presence of ad duplicates called for careful de-duplication, emphasizing the importance of thoughtful preprocessing to preserve meaningful content, such as unconventional word-number combinations or emojis, which might otherwise be overlooked. Selecting an effective NER tokenizer was pivotal; \emph{Longformer} emerged as our top choice, adept at handling the diverse data we encountered, including extracting and consolidating obvious metadata from ad descriptions. Post-processing steps, including normalizing entities like phone numbers for edge formation in the Relatedness Graph, were essential for creating meaningful connections between data points. This graph, although sparse, served as a critical backbone for our analysis, revealing interesting patterns and insights, as evidenced by Figure \ref{fig:cc_identifiers} and highlighting the skewed distribution of component sizes in our dataset (Table \ref{tab:cc_size}). Ultimately, the Relatedness Graph facilitated the pseudo-labeling process for OAD and HTRP, showcasing our research's intricate interplay of challenges.

\section{Conclusion}

Our research presents a methodology centered around the complexities of detecting organized crime, particularly in human trafficking. By integrating advanced NER techniques and utilizing the Longformer model for its adept handling of extensive texts, we have developed a dynamic dataset creation process that identifies non-trivial connections within ads. Our work emphasizes the importance of adaptable, privacy-aware approaches in dataset development, offering the research community a refined, scalable framework for navigating the initial challenges of data-centric investigations into organized crime.

\section*{Acknowledgements}

This work was funded by the National Science Foundation under grant CNS-2210091.

\bibliography{custom}

\appendix

\section{Ethics and Broader Impact Statement}

This research addresses the controversial and sensitive issue of detecting human trafficking within online commercial sex advertisements. Our primary goal is to identify linguistic traits that can help understand criminal communication in consumer-to-consumer online marketplaces. 

To protect potential victims of trafficking, we have chosen not to release the dataset. Instead, we provide a detailed protocol to allow reproducibility without compromising safety and privacy. This ensures that sensitive data is not exposed, minimizing the risk of harm to vulnerable individuals.

We acknowledge the potential misuse of our research, which could inadvertently target legitimate sex workers. To mitigate this risk, our findings only highlight patterns and indicators. It is crucial that any findings derived from our methodology be used as part of a broader, victim-centered approach prioritizing safety and well-being over punitive measures.

Our ethical considerations include:

\begin{itemize}
    \item \textbf{Victim Protection:} By withholding the dataset and focusing on methodological transparency, we prevent potential harm from the misuse of sensitive data.
    \item \textbf{Responsible Data Use:} We urge researchers and practitioners to collaborate with social scientists, legal experts, and victim advocacy groups to ensure ethical use of our protocol.
    \item \textbf{Contextual Analysis:} Our methodology should be used as a supplementary tool within a holistic investigative framework that includes qualitative assessments and corroborative evidence.
    \item \textbf{Stakeholder Impact:} We recognize the sensitive nature of our research and its potential impact on various stakeholders, including law enforcement, policymakers, researchers, and victims. Our goal is to contribute positively to combating human trafficking.
\end{itemize}

By prioritizing victim protection, promoting responsible data use, and encouraging a holistic approach to interpretation, we aim to make a meaningful contribution to the fight against human trafficking. Our commitment is to ensure that our work is used ethically and effectively to aid in identifying and protecting victims, while preventing misuse that could cause harm.

\end{document}